\documentclass[nohyperref]{article}

\usepackage{microtype}
\usepackage{graphicx}
\usepackage{subfigure}
\usepackage{booktabs} 

\usepackage{hyperref}

\usepackage{color,soul}

\newcommand{\image}{\mathbf{x}}
\newcommand{\ex}{\mathbf{e}}
\newcommand{\mask}{\mathbf{m}}

\usepackage[accepted]{icml2022}

\usepackage{amsmath}
\usepackage{amssymb}
\usepackage{mathtools}
\usepackage{amsthm}

\usepackage[capitalize,noabbrev]{cleveref}

\theoremstyle{plain}

\theoremstyle{definition}

\theoremstyle{remark}

\usepackage[textsize=tiny]{todonotes}

\icmltitlerunning{Psychological theory of XAI}

\begin{document}

\twocolumn[
\icmltitle{A Psychological Theory of Explainability}




\icmlsetsymbol{equal}{*}

\begin{icmlauthorlist}
\icmlauthor{Scott Cheng-Hsin Yang}{equal,rutgers}
\icmlauthor{Tomas Folke}{equal,rutgers}
\icmlauthor{Patrick Shafto}{rutgers,ias}
\end{icmlauthorlist}

\icmlaffiliation{rutgers}{Department of Mathematics and Computer Science, Rutgers University--Newark, New Jersey, USA}
\icmlaffiliation{ias}{School of Mathematics, Institute for Advanced Study, New Jersey, USA}

\icmlcorrespondingauthor{Scott Cheng-Hsin Yang}{scott.cheng.hsin.yang@gmail.com}
\icmlcorrespondingauthor{Tomas Folke}{tomas.folke@gmail.com}

\icmlkeywords{Machine Learning, ICML}

\vskip 0.3in
]



\printAffiliationsAndNotice{\icmlEqualContribution} 

\begin{abstract}
The goal of explainable Artificial Intelligence (XAI) is to generate human-interpretable explanations, but there are no computationally precise theories of how humans interpret AI generated explanations. The lack of theory means that validation of XAI must be done empirically, on a case-by-case basis, which prevents systematic theory-building in XAI. We propose a psychological theory of how humans draw conclusions from saliency maps, the most common form of XAI explanation, which for the first time allows for precise prediction of explainee inference conditioned on explanation. Our theory posits that absent explanation humans expect the AI to make similar decisions to themselves, and that they interpret an explanation by comparison to the explanations they themselves would give. Comparison is formalized via Shepard's universal law of generalization in a similarity space, a classic theory from cognitive science. A pre-registered user study on AI image classifications with saliency map explanations demonstrate that our theory quantitatively matches participants' predictions of the AI.
\end{abstract}

\section{Introduction}

Modern AI systems are applied to high-impact domains, such as medicine \cite{pesapane2018_ai_radiology} and finance \cite{gogas2021machine}. 
These systems, powered by deep neural networks, are notoriously opaque \cite{adadi2018peeking}, making supervision and safe deployment challenging \cite{guidotti2018survey}. 
The field of explainable AI has produced many techniques for improving the legibility of AI decisions to human users and regulators \cite{gunning2019darpa,arrieta2020explainable}.
XAI has focused on developing new methods that show high performance on technical metrics related to faithfulness \cite{hooker2019benchmark, yeh2019fidelity, sundararajan2017axiomatic} and explanation complexity \cite{blanc2021provably, ribeiro2016lime, qi2020visualizing}, but there is no way to assess which methods will do well for a given use-case \cite{doshi2017towards}. In other words, there is no theory of explainability. 

The goal of XAI is for humans to understand a target AI system \cite{gunning2019darpa, miller2019explanation, doshi2017towards}. This ``understanding" can be formalized as congruence between the AI's input-output mapping, and the human mental model of that mapping.
Good explanations shift the human mental model to achieve this congruence.
As a consequence, a theory of explainability can be naturally formalized as Bayesian updating. The initial human (mis-) conception of the AI serves as a prior, and the explanations provided modify this prior via a likelihood function that captures human inferential processes. 
An alternative approach to explainability would be to estimate the relationship between explanation and human inference as a ML problem, for example, by training on saliency maps as input, and human inference of AI classification as output. 
However, such an approach does not model human inference explicitly, and as a result suffers from all the standard problems with black box models, such as unknown generalization properties, fragility to out-of-distribution observations, and challenges relating to debugging and auditing. 


To avoid building black-boxes to explain black-boxes, a formal theory of explainability  must be constrained by cognitive science.
We propose that humans model AI systems the same way they model any other agent, allowing us to draw on psychological work on belief-formation, generalization, and theory-of-mind.
Our theory states that people project their own beliefs onto the AI and update their beliefs based on how they generalize self-generated explanations to XAI explanations in a similarity space \cite{yang2021mitigating,shepard1987toward,sloman1998similarity}. 
We built a cognitive model formalizing these ideas, and compared its predictions to human inference in a user study. We asked users to infer AI classification on images given saliency-map explanations, and found that our image-level model predictions correlated strongly with user responses (Spearman's $\rho = .86$). 
Variations to the similarity space or the generalization law that reduced the theory's psychological plausibility also harmed the model's predictive performance. Our theory quantitatively predicts human inferences from explanation, and can thus guide the development and deployment of explainable AI techniques.
Furthermore, our results show that well-established constructs from cognitive science offer realistic user mental models for explainable AI, illustrating the value of bridging the two fields.

\section{Related work}

\textbf{Calls for human-centered explainability.} There has been a proliferation of survey papers of XAI techniques in recent years that have attempted to synthesize existing knowledge and suggest taxonomies of XAI methods (see \citet{linardatos2021explainable} for a recent example that lists and discusses most previous surveys). Some of the most influential surveys and position papers put human understanding at the core of their definition of explainability \cite{miller2019explanation, doshi2017towards, adadi2018peeking}, and bemoan the lack of mathematical formalization.
In contrast, many attempts to formalize desiderata for XAI have focused on faithfulness, which concerns how accurately the explanation captures the behavior of the AI system to be explained \cite{hooker2019benchmark, yeh2019fidelity,li2021experimental, sundararajan2017axiomatic}, rather than human interpretability. 
In the absence of a formal theory, interpretability can only be evaluated by user studies \cite{hase2020evaluating, jesus2021can, adebayo2020debugging}, which are expensive and challenging to run.
Because of the range of applications and the speed at which new XAI methods emerge, naive empiricism is insufficient to determine which XAI method is most appropriate for a specific problem. Theory is necessary to determine to what extent empirical findings generalize.

\textbf{Practical issues with explanations.} Empirical studies have shown that practitioners can over-interpret explanations to shepherd false beliefs or under-use explanations because of lack of contextualization \cite{kaur2020interpreting}. Furthermore, explanations from different methods often disagree \cite{krishna2022disagreement}, and some explanations might mislead experts,leading to mistakes and miscalibrated trust \cite{lakkaraju2020fool}. All these issues emphasize the importance of having a quantitative theory of how humans interpret explanations.

\textbf{Interpretability and explanation sparsity.} When researchers do consider the interpretability of explanations, they often treat it as synonymous with the sparsity of the explanation \cite{blanc2021provably, ribeiro2016lime, qi2020visualizing}.
While it is generally true that sparse explanations are typically preferable to more complex ones, it is a mistake to only consider sparsity when attempting to achieve interpretability.
First, sparsity comes in many forms, and its impact on interpretability may vary depending on the users' expertise, explanation presentation, and the task. While some studies have attempted to investigate the impact of various forms of sparsity on interpretability \cite{narayanan2018humans, poursabzi2021manipulating}, the domain is too large for exhaustive experimentation. 
Second, there is often a trade-off between the sparsity of an explanation and its expressivity, so penalizing explanation complexity is not always beneficial \cite{doshi2017towards}.
Most importantly, an exclusive focus on sparsity ignores any inferential biases of the explainee, which is essential to how explanations are understood.
An explicit model of explainee inference subsumes sparsity by targeting the quantity we truly care about: what human infers about the AI's input-output mapping based on the explanations.
The lack of explicit models of explainee inference results in uncertainty about the effectiveness of XAI methods in the context of new domains. 

\textbf{Black-box models of inference from explanations.} There has been some attempts to approximate the impact of explanations on human users, by simulating the user with black-box statistical techniques \cite{pruthi2022evaluating, chen2021simulated, yang2021mitigating}. \citet{chen2021simulated} provide a sophisticated example of this approach. They used supervised learning to teach an explainee model to associate explanations with labelled outcomes of interest, and then tested the explainee model on novel explanations generated by the same method. 
However, they only qualitatively compared their simulated results to actual user data (collected by other labs). 
Furthermore, even if they had shown that their explainee model predictions quantitatively matched human explainee inference, their model is itself a black-box, so it could not be used to inform better explanation generation, and the model's generalization properties to new contexts is uncertain.
\citet{yang2021mitigating} argued for explainee models informed by human psychology, but still used a neural net to model the explainee. Their empirical results also showed that their explainee model qualitatively predicted human judgements, but their explainee predictions only weakly correlate with the human data. Interestingly, while they argued that humans model the AI the same way as they model other agents, the authors did not implement these ideas in their formal model of the explainee.

\section{Theory}

\subsection{Hypotheses}
Here we introduce a quantitative theory of user inference of AI decisions from explanations. Our theory states that people project their own beliefs onto the AI, and update beliefs based on explanation via generalization in similarity space in order to predict AI behavior. Specifically, the belief updating involves comparing the observed explanation to the \textit{projected} explanation a person would give to justify a given judgement. 

Theories of social cognition and neuroscience suggest that we model other agents by simulation \cite{gallese1998mirror, buckner2007self, schurz2021toward}, and empirical evidence suggests that if we lack specific information about a person's preferences or beliefs, our initial assumption is that they will be similar to us \cite{tarantola2017prior, suzuki2016behavioral, harris2018accounting}. Evidence from human-computer interaction suggests that people treat computers as social agents \cite{nass2000machines}, even attributing them personality \cite{nass1995can}.
For these reasons, we hypothesize that human users will not model the AI as a completely unknown entity, but will rather project their own beliefs onto the AI system (Hypothesis 1). Successful explanations should inform this belief-projection so as to improve the fidelity between user beliefs and AI behavior when belief-projection is misleading compared to when it is not (Hypothesis 2) \cite{yang2021mitigating}. Throughout the paper, fidelity refers to the probability that the human's prediction of AI classification matches the AI's actual classification. The effect of an explanation on human beliefs depends on how well projected explanations match observed explanations, which is hypothesized to be quantitatively predicted by generalization in feature similarity space (Hypothesis 3). 

A successful explanation updates the explainee's belief of AI behavior towards the actual AI behavior. As such, inference from explanation is naturally modelled by Bayes' rule. The Bayesian formulation puts a natural minimum criterion that any successful theory of explanation needs to exceed: a successful cognitive model should capture humans' beliefs post-exposure to explanation better than a model that only relies on their prior beliefs. 
We hypothesize that our theory with a psychologically informed likelihood will match human inference from explanation better than a prior-only model that is based on human beliefs about AI classifications when no explanations are presented (Hypothesis  4).
 
The quantitative degree of belief update depends on generalization in similarity space \cite{goldstone2012similarity}. 
The comparison between projected and observed explanations is assessed in similarity space. There has been extensive work on the mathematical form of psychological similarity judgements \cite{goldstone2012similarity}. 
We use a symmetric, feature-based similarity measure proposed by Sloman \cite{sloman1998similarity}. To evaluate whether psychological plausibility of the theory's similarity space impacts its predictive performance, we contrast the Sloman similarity to L1 norm --- a raw distance metric commonly used in computer science. 
We hypothesize that a model that compares projected and observed explanations in a psychologically natural similarity space will match human beliefs better than L1 distance would (Hypothesis 5). 

Shepard showed that generalization between stimuli in similarity space follows a monotonically decreasing, approximately exponential, function \cite{shepard1987toward}.  
To test whether this law of generalization holds for inference from explanation, we contrast a model with an exponential likelihood to one with a beta-distribution where the parameters are constrained to violate monotonicity. 
We hypothesize that the monotonically decaying likelihood will capture human beliefs better than the non-monotonic alternative would (Hypothesis 6).

\subsection{Formalism}\label{sec:formalism}
Let $c$ be a class of interest, and $\image$ be an input image with $W \times H$ pixels. The black-box AI model to be explained, $f(\image,c)$, takes a pair of $\image$ and $c$ as inputs, and outputs the probability that $x$ is an instance of $c$. A saliency-map XAI method, $g(\image,c,f)$, takes the triplet $\image$, $c$ and $f$ as inputs, and outputs an explanation $\ex \in [0,1]^{W \times H}$ that is the same size as $\image$. Each pixel of the saliency map explanation $\ex$ represents the importance of that pixel in $\image$ to $f$'s classification of $\image$ as $c$. The main quantity we set out to model is the human explainee's inference about the AI's output given the $\image$, $c$, $\ex$, and the alternative class(es) that are involved, which we denote as $P(c \mid \ex,\image)$. Cast in a Bayesian framework, the explainee's inference can be expressed as a posterior:
\begin{equation}
P(c \mid \ex,\image) = \frac{p(\ex \mid c,\image)P(c \mid \image)}
{\sum_{c'}p(\ex \mid c',\image)P(c' \mid \image)}.
\label{eq:posterior}
\end{equation}
The prior $P(c \mid \image)$ is the explainee's inference of the AI's classification without any explanation. The likelihood $p(\ex \mid c,\image)$ is the probability that the explainee themself would provide the observed saliency map $\ex$ as the explanation for assigning class $c$ to image $\image$. The sum over $c'$ includes the class of interest and the alternative class(es) in contrast.

Figure~\ref{fig:intuition} depicts how the theory is constructed and validated. First, we instantiate the theory in a two-forced alternative choice (2AFC) task, limiting the alternative class to a single foil class. Using such as task, we measured the prior in a control condition where participants inferred the AI's classification without any explanation. Quantitatively, the prior $P(c \mid \image)$ is set to be the average response across participants in the control condition for image $\image$. The likelihood is constructed from a set of drawing experiments that measured the explanations that the participants themselves would generate. These participant-generated explanations, $\ex^P$, are then compared to the observed XAI-generated explanation $\ex$ in a similarity space to produce the likelihood $p(\ex \mid c,\image)$. Finally, the prior and likelihoods are combined using Bayes' rule to output the posterior $P(c \mid \ex,\image)$. On an image-by-image base, the posterior is then compared against its experimental measurements, $P^*(c \mid \ex,\image)$, in an explanation condition, where participants' inferred the AI's classification with the XAI-generated explanation $\ex$.

\textbf{Likelihood construction.}
The intuition behind the construction of the likelihood is that humans interpret the observed explanation by comparing it to the explanations that they themselves would generate. If the observed explanation is similar to the explanation that they would generate for a particular class, the explanation will push the explainees' inference to favor that class. In order to quantify this intuition, we formalized this comparison with the law of generalization in a psychologically plausible similarity space.

Following Shepard's universal law of generalization \cite{shepard1987toward}, we expect the likelihood to decay as a function of dissimilarity between the observed saliency map $\ex$ and the projected saliency map $\ex^{P}$ from the drawing experiment. For simplicity, we use the celebrated exponential form:
\begin{equation}
p(\ex \mid c,\image) = \lambda \exp\!\left[ -\lambda\, \left(1-sim\!\left[ \ex(c,\image) \,,\,\ex^{P}\!(c,\image) \right] \right) \right],
\label{eq:likelihood}
\end{equation}
where $\lambda>0$ is the only free parameter in our cognitive model, used to calibrate the rate at which generalization decays with dissimilarity. The saliency maps $\ex$ and $\ex^P$ are expressed as a function of $c$ and $\image$ to clarify where the class and image contribute to the equation. Following Sloman \cite{sloman1998similarity}, we adopt a simple parameter-free form of similarity (cosine similarity) given by 
\begin{align}
sim\!\left[\ex(c,\image) \,,\, \ex^{P}\!(c,\image) \right] = \frac{\sum_i \ex_i(c,\image)\, \ex^{P}_i\!(c,\image)}
{ \sqrt{ \sum_i {\ex_i(c,\image)}^2} \sqrt{ \sum_i {\ex^{P}_i\!(c,\image)}^2} },
\label{eq:similarity}
\end{align}
where the $i$ index the pixel of the saliency maps. See Appendix~\ref{app:sim} for details on the similarity calculations.

\textbf{Ablation models.}
In order to evaluate whether the likelihood in the full cognitive model successfully captures belief-updating in response to specific observed saliency maps, we create a series of ablation models in which the likelihood deviates from psychological plausibility. The first ablation model is a \textbf{prior-only model} that excludes the likelihood term, such that the posterior is equal to the prior. The second model is an \textbf{L1-norm model} that replaces Sloman similarity in Equation~\ref{eq:similarity} with the L1-norm distance $\sum_i \lvert \ex_i(c,\image) - \ex^{P}_i\!(c,\image) \rvert$. The L1-norm model evaluates the impact of using a pixel-based comparison in L1 norm, which is believed to be less natural for humans than a feature-based comparison captured by the Sloman similarity. Lastly, the third model is a \textbf{Beta-distribution model} in which the exponential distribution of Equation~\ref{eq:likelihood} is replaced with a Beta distribution where its two shape parameters are set to equal each other. Shepard showed that generalization follows a monotonic decreasing function \cite{shepard1987toward}; thus, the Beta-distribution model tests whether this law of generalization holds when linking participants' similarity judgements to their specific responses. The Beta distribution takes the form:
\begin{figure}[H]
  \centering
  \includegraphics[width=\columnwidth]{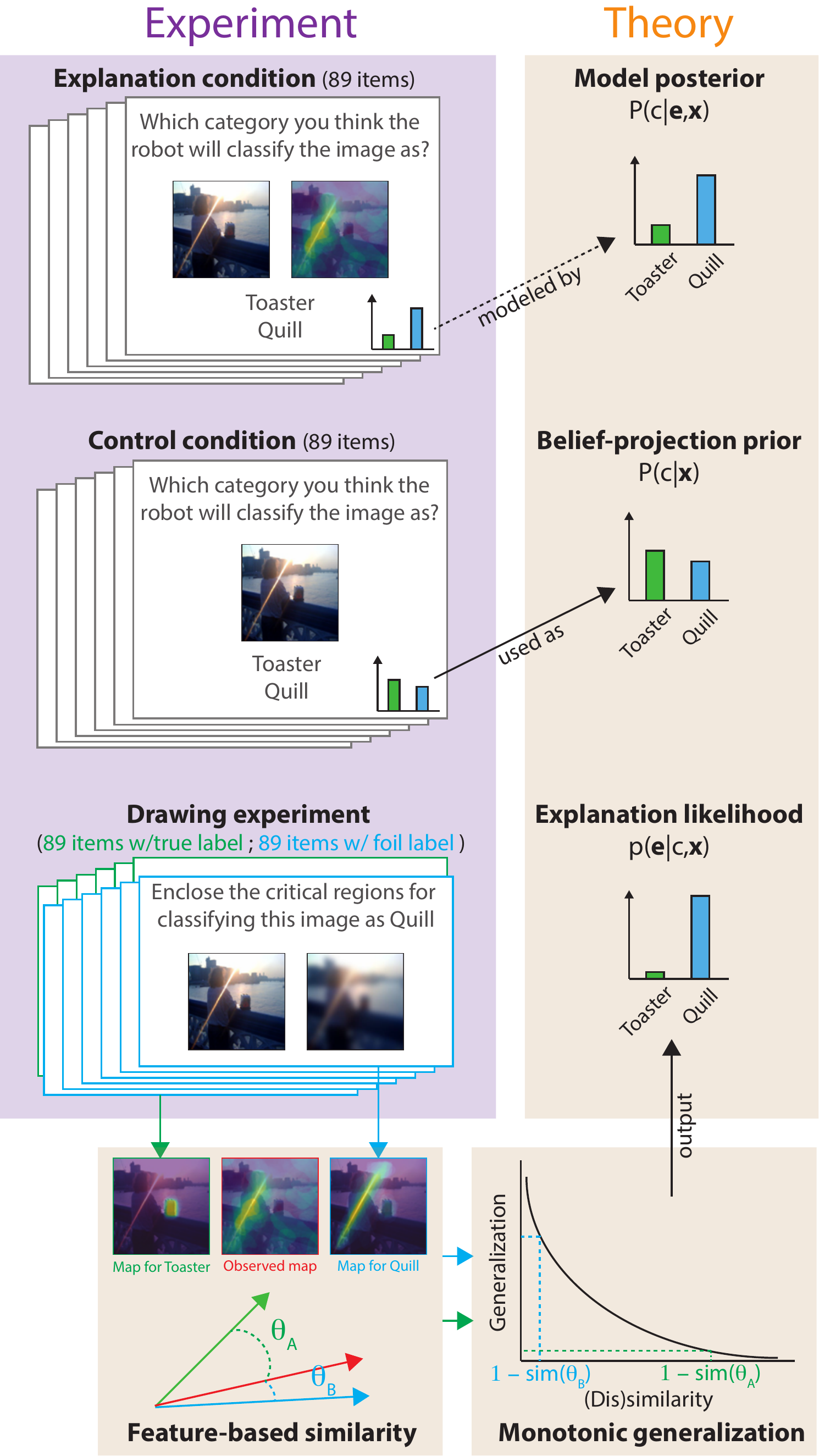}
  \caption{The relationship between the theory and experiments. Human interpretation of an explanation, $P^*(c \mid \ex, \image)$, is measured by the participants' responses when viewing a saliency map explanation (the classification experiment's  explanation condition) and modeled by the theory's posterior, $P(c \mid \ex, \image)$. The participants' responses to the same stimuli absent explanation (the classification experiment's control condition) are taken to be the belief-projection prior of the cognitive model, $P(c \mid \image)$. Different participants enclosed important regions of the same images, contingent on a given class (the drawing experiment). The regions of interest recorded in the drawing experiment are used to compute the explanation likelihood, $p(\ex \mid c, \image)$. The computation involves calculating how well the XAI-generated saliency map generalizes to the average participant-generated saliency maps in the feature-based similarity space. The figure illustrates a trial in which the explanation helped participants to shift their belief from favoring that the AI classified the image as Toaster to a strong and correct belief that the AI classified the image as Quill.}
  \label{fig:intuition}
\end{figure}
\begin{align}
\begin{split}
p(\ex \mid c,\image) = \frac{\Gamma(2\lambda)}{\Gamma(\lambda)\Gamma(\lambda)}
sim\!\left[ \ex(c,\image) \,,\,\ex^{P}\!(c,\image) \right]^{\lambda-1} \\
\times \left( 1-sim\!\left[ \ex(c,\image) \,,\,\ex^{P}\!(c,\image) \right] \right)^{\lambda-1} 
,
\end{split}\label{eq:beta}
\end{align}
where $\Gamma(\cdot)$ is the gamma function, and $\lambda>0$ is a free shape parameter. By restricting the two parameters of the Beta distribution to be equal ($\alpha = \beta = \lambda$), we force the likelihood to violate monotonicity and thus contrast against the consistently decaying exponential form. In Section~\ref{sec:statistical_analysis} we detail how the hypotheses relate to the quantities introduced in this section.

\section{Experiments}
We tested our theory and formalized cognitive model in the domain of image classification. Participants saw an image $\image$ and were asked to report which class $c$ they believed the AI would assign to $\image$ in a two-alternative-forced-choice task (see Figure~\ref{fig:intuition}). Each choice was between the ground-truth label of the image, and a foil label that was either the AI's classification (when the AI made a mistake) or the second-most likely category according to the AI (when the AI made a correct classification). 
Participants in this \textbf{classification experiment} were randomly assigned to one of two conditions: in the control condition they inferred the AI classification without seeing an explanation; in the explanation condition they made the same judgement but were also exposed to a saliency map explanation $\ex$ that highlights regions of $\image$ that highly influence the AI's decision \cite{yang2021mitigating, petsiuk2018rise}.
To estimate the projected saliency maps contingent on a specific classification $\ex^{P}\!(c,\image)$, we ran a \textbf{drawing experiment}. Participants were asked to enclose which regions of the image they thought were important for a given classification. This drawing experiment involved two between-subject conditions: one for the true labels and one for the foil labels. The average of these human-generated regions were taken to be the projected saliency maps that humans would use to interpret the observed saliency maps.

The classification experiment and the drawing experiment shared the same 89 images, with one image shown per trial. The AI, a ResNet-50 model \cite{he2016deep}, correctly classified 30 images and misclassified the remaining 59.

\subsection{Classification Experiment}

\textbf{Task structure.} Participants completed a 2AFC where they had to predict how the AI classified an image. The experiment consisted of two between-participant conditions: a \textit{control} condition without explanation and an \textit{explanation} condition with saliency map explanation.

\textbf{Image selection.} Participants viewed 89 images. These images belonged to three distinct types all sharing the same classes. The types were correctly classified images drawn from ImageNet \cite{ILSVRC15}, misclassified images drawn from ImageNet, and misclassified images drawn from the Natural Adversarial ImageNet dataset \cite{hendrycks2021natural}. The latter contains a 200-class subset of the original ImageNet 1000 classes. From these 200 classes, we selected 30 classes that span the spectrum of model accuracy based on the predictions of ResNet-50 on the normal ImageNet's validation set. We selected one image from each class, for each of our three image types. All images were randomly sampled from the chosen classes and datasets. In our pilot work, we found no clear distinctions between AI mistakes based on the adversarial images and the AI mistake based on the standard ImageNet images, so in this study we treat all mistake trials as a single type.

To form the 2AFC trials, we selected two possible labels for each image. One of these labels was always the ground truth label of the image, whereas the other was a foil. For correctly classified images, the foil was the class that ResNet-50 found most confusable according to the confusion matrix constructed on ResNet-50's predictions on normal ImageNet's validation set. For misclassified images, the foil was the class that ResNet misclassified the image as.

The above procedure produces 90 trials that specify the image, the ground-truth class, and the foil class. The original experiment that inspired this classification experiment involved both example images and saliency maps as explanations. 
One of the trials was excluded because no good explanatory examples were found, leaving a total of 89 trials. 
Since the symmetry of trial types is not central to the hypotheses we test in this study, we adopt that setup as is.

\textbf{Saliency map generation.} We used the method of Bayesian Teaching to generate saliency map explanation \cite{yang2021mitigating}. The saliency value of each pixel of the explanation is related to the probability that the corresponding pixel in the image helps the AI to arrive at the targeted prediction. See Appendix~\ref{app:BTmap} for technical details.

\textbf{Participants.} The study protocol was approved by Rutgers University IRB. All research was performed in accordance with the approved study protocol. An IRB-approved consent page was displayed before the experiment. Informed consent was obtained from all participants prior to the start of the experiment. Our preregistered aim was to test participants until we had at least 40 participants per condition who passed our inclusion criteria. For the control condition we tested 50 participants and excluded 9, leaving us with a final sample of 41 participants. For the explanation condition, we tested 59 participants and excluded 13, leaving us with a final sample of 46.

\textbf{Exclusion criteria.} In line with our preregistration, we excluded participants based on total response time. We had different thresholds for different conditions since the explanation condition involved integrating more information than the control condition.
For the control condition we excluded any participant that had a total response time less than three minutes, corresponding to approximately 2 seconds per item.
For the explanation condition we excluded participants that took less than five minutes to complete the experiment, corresponding to an average response time of 3.4 seconds per item. 

\subsection{Drawing Experiment}

\textbf{Task structure.} Participants were shown the 89 images from the classification experiment. They generated a mask for each image from that experiment, by drawing on a blurred version of the image to indicate what regions of the image was important for a given classification. The experiment consisted of two between-participant conditions depending on which class participants were asked to explain (ground-truth or foil). In the ground-truth condition, these instructions read: ``enclose \textsc{category\_name}"; in the foil condition these instructions read: ``enclose the critical regions you believe the robot attended to when determining the image contains \textsc{category\_name}".

\textbf{Participants.} The study protocol has been approved under the same IRB used for the classification experiment. Our aim was to include data from at least 40 participants per condition post exclusions. Because we had high attrition rates during our pilot, we tested 160 participants participants in the ground-truth condition and 80 participants in the foil condition. We excluded 95 from the ground-truth condition and 8 from the foil condition, leading to ultimate sample sizes of 65 and 72, respectively.

\textbf{Exclusion criterion.} In line with our preregistration, we excluded participants based on how much their masks deviated from the aggregate image-level masks across participants. The intuition behind this (validated in our pilot) is that people who pay attention to the image tend to report that the same regions as important, whereas people who do not pay attention deviate from this consensus. We quantified how much participants deviate from the consensus on an image by the following steps: First, for each participant compute the L2 norm between the aggregate and the participant's specific mask. Second, compute the participant-level mean of these L2 norms. Third, compute a Z-score for each participant using the participant-wise mean and assuming a half-normal distribution. 
Any participant with a Z-score greater than 1.5 is excluded for that image. The L2 norm, the participant-wise means, and the Z-scores are then recomputed on the post-exclusion data, and this process is repeated until all participants have a Z-score less than or equal to 1.5. We arrived at this threshold from exploratory analysis of our pilot data.

\textbf{Preregistration.} The classification experiment, the drawing experiment, the mathematical models, and our hypothesis tests were all preregistered prior to collecting the data for the results presented in the paper 
(link to pre-registration to be provided upon acceptance or reviewer request; the preregistration document and our analysis code has been anonymized and uploaded in the supplementary folder).
We only deviated from our preregistration in one respect: we used a more generous inclusion criterion for the drawing experiment. Using our preregistered threshold reduced our sample size, but did not otherwise impact our results, see Appendix~\ref{app:exclusions} for details.

\subsection{Model evaluation}

\textbf{Fitting model parameter.} To fully specify the models introduced in Section~\ref{sec:formalism}, we need to assign values to $\lambda$ in Equations~\ref{eq:likelihood} and ~\ref{eq:beta}. We select $\lambda$ values for these models by minimizing the sum of the squared error. The error term was computed on the image-level between the model posterior probabilities and the empirical responses from the explanation condition in the classification experiment.

\textbf{Leave-one-out cross-validation.} For the final three hypotheses, we expected that a cognitive model with psychologically informed components --- prior belief-projection, feature-based similarity, and monotonic generalization --- would match human responses in the explanation condition better than alternatives would. To compare the predictive performance of the full model to the alternatives, we used leave-one-out cross-validation (LOO-CV) to control for model complexity. For cognitive models with a $\lambda$ parameter, we fitted the model on 88 of the 89 trials, and used the obtained $\lambda$ to compute the squared error between the fitted model and the remaining data point.
For each left-out item, we computed the squared error between the model's posterior and the empirical response from the explanation condition in the classification experiment (LOO-CV MSE). We then used paired t-tests to evaluate if the LOO-CV MSE was significantly lower for the full model than the ablated alternatives.

\subsection{Statistical analysis strategy}\label{sec:statistical_analysis}
\textbf{Hypothesis 1.}
Our first hypothesis was that participants' prior belief of the AI classification, $P(c \mid \image)$, would not be uniform,
but that the participants would expect the AI to make similar classifications to themselves. Because humans tend to find this type of image classification easy, the result of this belief projection would be that participants in the control condition of the classification experiment would pick the ground truth label more often than would be expected by chance \cite{yang2021mitigating}. We tested this hypothesis with chi-squared test on the proportions of participant responses that matched the ground truth.

\textbf{Hypothesis 2.}
Our second hypothesis was that successful explanations would update the belief-projection prior by highlighting the discrepancy between what features participants expected the AI to attend to and the features it actually found important. Because this discrepancy is generally larger when the AI makes a mistake, we hypothesized that explanations would help participants identify AI mistakes. However, sometimes the AI will attend to strange features even when it is correct, meaning that explanations would be less helpful (or possibly harmful) for identifying correct AI classifications. 

To evaluate this hypothesis, we computed the average image-level fidelity between the participants' predictions of the AI and the AI's actual classifications for each of the experimental conditions, resulting in 178 observations. 
In terms of the quantities introduced in Section~\ref{sec:formalism}, the fidelity in the control and explanations conditions are $P(c=c_{AI} \mid \image)$ and $P^*(c=c_{AI} \mid \ex, \image)$, respectively, where $c_{AI}$ is the AI's classification of $\image$.
We then performed statistical analysis on the image-level fidelity using linear regression:
\begin{align}
Fidelity \sim \;&Normal(\mu_{i}, \sigma^{2}) \nonumber\\
\mu_{i} = \; &\beta_{0} + \beta_{1}A_i + \beta_{2}E_i
+ \beta_{3}(A_i \times E_i).
\label{eq:lm}
\end{align}
Here, $A_i$ is a binary independent variable coded as 1 if the AI correctly classified the image of that trial, and 0 otherwise. $E_i$ is another binary independent variable coded as 1 if the response belonged to the explanation condition, and 0 if it belonged to the control condition. $i$ corresponds to a single image in one of the conditions and ranges from $1$ to $178$, which is the total number of images for both conditions.

\textbf{Hypothesis 3.}
Our third hypothesis was that our cognitive model would capture the change in fidelity from the control to the explanation condition, depending on AI correctness as outlined for hypothesis 2, despite the model being blind to whether the AI made a mistake on a given trial. 

To evaluate whether our cognitive model could qualitatively recover the empirical patterns, we ran a variation of the regression analysis for Hypothesis 2, where we replaced the fidelity of participant choices with the fidelity of our cognitive model's predictions for the explanation trials; that is, we analyzed the model-generated $P(c=c_{AI} \mid \ex, \image)$ instead of the measured $P^*(c=c_{AI} \mid \ex, \image)$. 

\textbf{Hypothesis 4.}
Our fourth hypothesis was that the likelihood in our cognitive model captures belief-updating from specific explanations, meaning that the full model should reliably outperform the prior-only model. 
Performance is captured by the match between model predictions and human responses in the explanation condition. That is, we expect the MSE between $P^*(c \mid \ex, \image)$ and $P(c \mid \ex, \image)$ to be smaller than that between $P^*(c \mid \ex, \image)$ and $P(c \mid \image)$.

\textbf{Hypothesis 5.}
Our fifth hypothesis was that our full model, which used symmetric Sloman similarity, would match human responses better than the L1-norm model would.
That is, we expect the MSE between $P^*(c \mid \ex, \image)$ and $P(c \mid \ex, \image)$ to be smaller than that between $P^*(c \mid \ex, \image)$ and $P_\mathrm{L1}(c \mid \ex, \image)$, where $P_\mathrm{L1}(c \mid \ex, \image)$ denotes the posterior obtained from the L1-norm model.
In this context, L1-norm captures dissimilarity as the sum of the unsigned pixel-wise differences, which serves as a good foil since pixels are too granular to be psychologically meaningful features. 

\textbf{Hypothesis 6.}
Finally, our sixth hypothesis was that translating the similarity judgement between the observed explanation and the projected explanation to a response about AI classification is determined by the probability that the response to the projected explanation would generalize to the observed explanation. Shepard compellingly demonstrated that generalization decays monotonically, so we hypothesized that a generalization distribution that obeys this law would capture the human responses better than a distribution that does not. To formally test this, we compared our full model --- that used an exponential distribution --- to the Beta-distribution model that used a likelihood that was constrained to violate monotonicity. 
We expect the MSE between $P^*(c \mid \ex, \image)$ and $P(c \mid \ex, \image)$ to be smaller than that between $P^*(c \mid \ex, \image)$ and $P_\mathrm{Beta}(c \mid \ex, \image)$, where $P_\mathrm{Beta}(c \mid \ex, \image)$ denotes the posterior from the Beta-disribution model.

\textbf{Not preregistered analysis.}
Though not a preregistered hypothesis, we wanted to test how well the model posteriors $P(c \mid \ex, \image)$ matched the empirical data $P^*(c \mid \ex, \image)$ from the explanation condition on an image-by-image level. We tested this with Spearman correlation between the fidelity of the empirical data and the fidelity of the model posteriors.

\section{Results}

\begin{figure}[!t]
  \centering
  \includegraphics[width=\columnwidth]{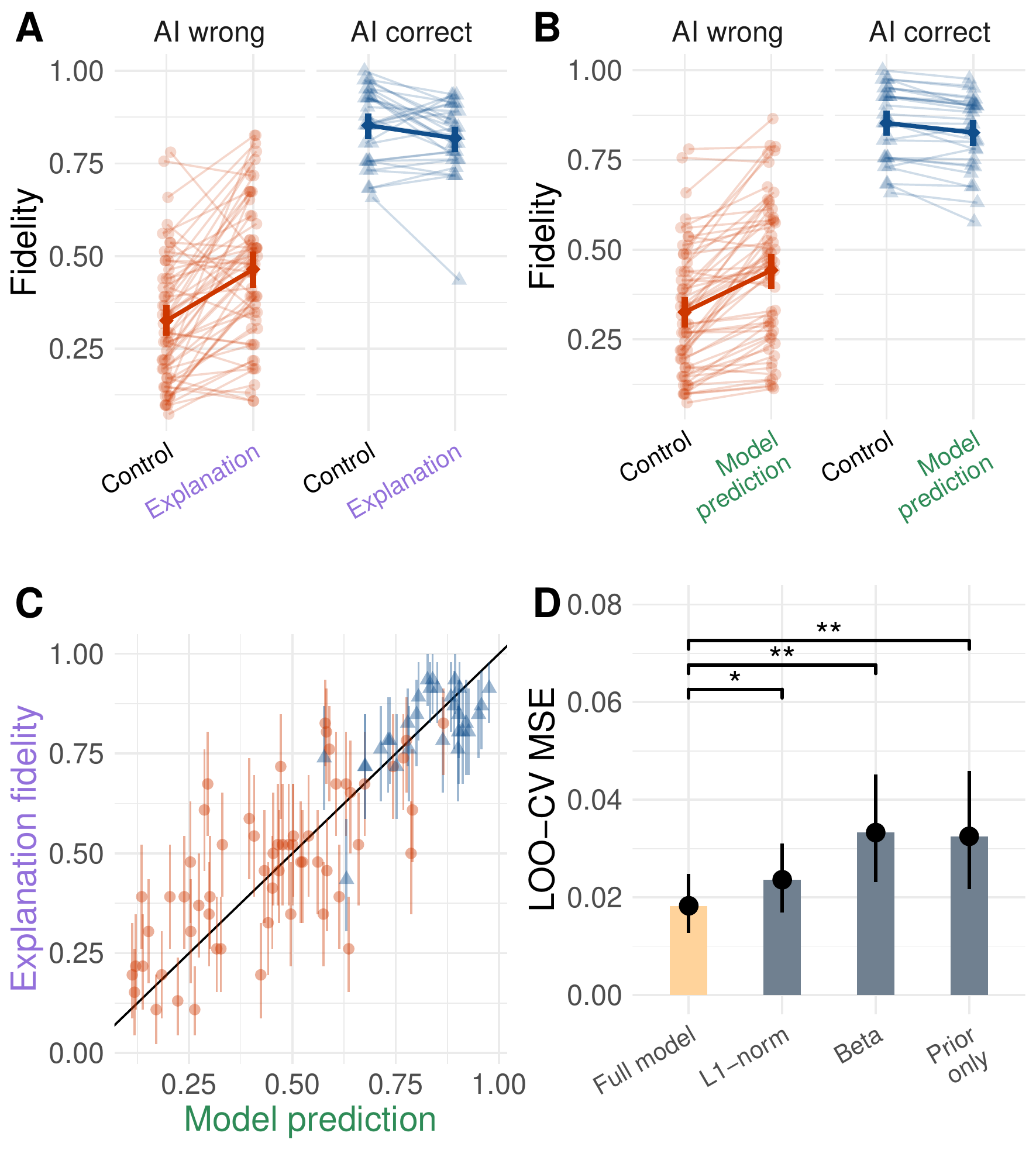}
  \caption{\textbf{(A)} Experimental results from the classification experiment. Explanations increase the fidelity between participant responses and AI classifications on trials when the AI is wrong, but slightly decrease fidelity when the AI is correct. Semi-transparent points show image-level data; solid points show condition-level means. \textbf{(B)} Model predictions of participant responses. Our cognitive model recovers the qualitative patterns of the empirical data. \textbf{(C)} Scatter plot between the empirical image-level fidelity and our model predictions. \textbf{(D)} Leave-One-Out Cross-Validation MSE for our model compared to ablated alternatives. All error bars show bootstrapped 95\% confidence intervals.} 
  \label{fig:results}
\end{figure}

All of our preregistered hypotheses were supported (see Appendix \ref{app:results} for details). First, absent explanation participants responded that the AI would correctly classify the image in 73\% of the trials ($\chi^{2}$ = 802.28, $p$ $<$ .0001), which is consistent with belief-projection because it implies that participants expected the AI to get most trials right in a task that they themselves find easy. Second, explanations improved the fidelity between participant responses and AI classifications when the AI makes a mistake ($\beta$ = 0.14, SE = 0.03, $t$ = 4.69, $p$ $<$ .0001; see Figure~\ref{fig:results}A), and the impact of explanations on fidelity was reduced when the AI is correct ($\beta$ = $-$0.17, SE = 0.05, $t$ = $-$3.40, $p$ $<$ .001). 
Third, our model predictions qualitatively match the empirical data, as our cognitive model also predicts that explanations will increase fidelity on mistake trials ($\beta$ = 0.12, SE = 0.03, $t$ = 3.90, $p$ $<$ .001), and that the impact of explanations should be less pronounced when the AI is correct ($\beta$ = $-$0.14, SE = 0.05, $t$ = $-$2.77, $p$ = .006; see Figure~\ref{fig:results}B). To obtain a general estimate of the model effectiveness in predicting human judgments, we ran a (not preregistered) Spearman correlation between fidelity based on the empirical data and fidelity based on the model predictions, which was statistically significant (Spearman's $\rho$ = .86, $p$ $<$ .0001; see Figure~\ref{fig:results}C). 

\begin{figure}[!t]
  \centering
  \includegraphics[width=\columnwidth]{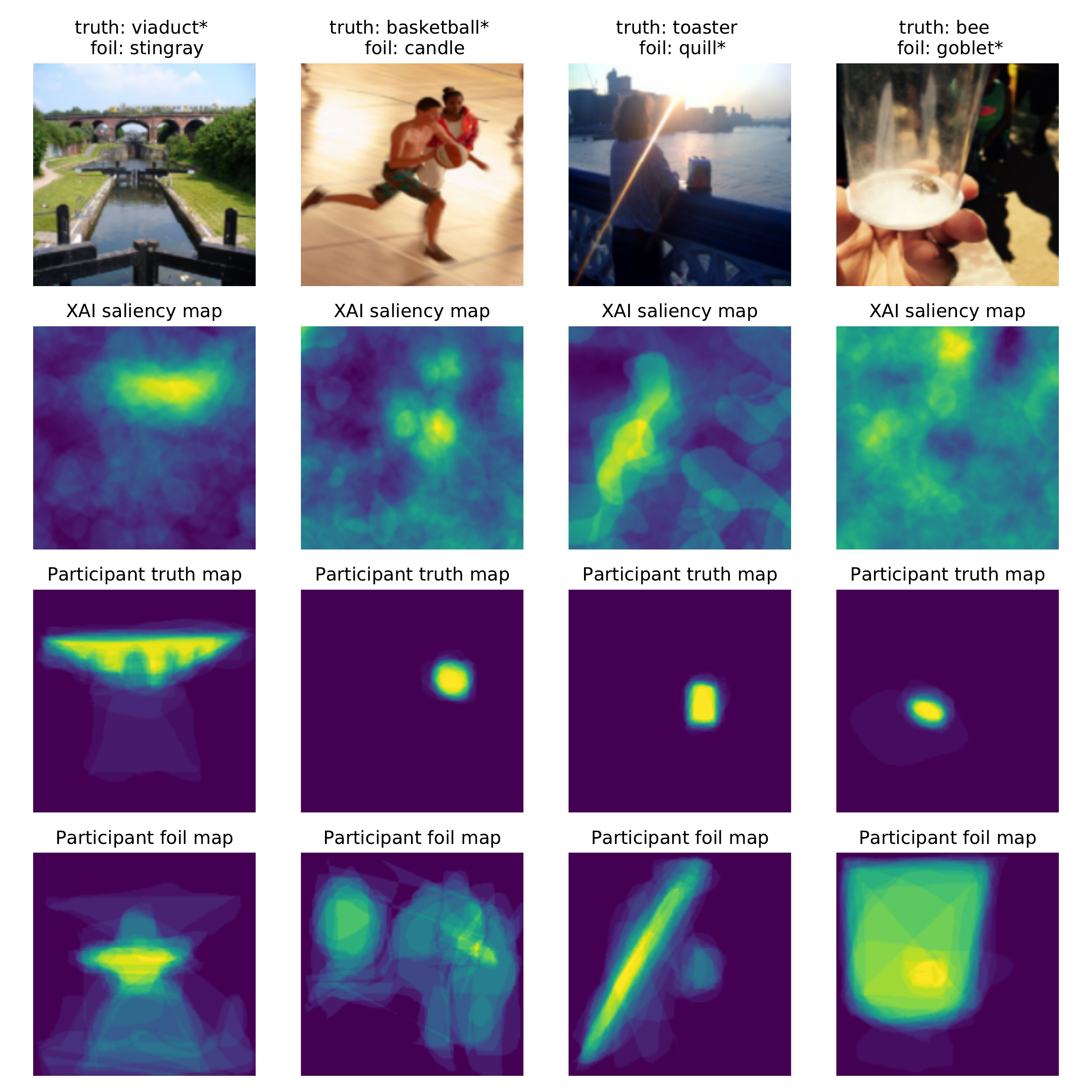}
  \caption{Examples of XAI saliency maps and participant drawn maps. Title of each column show the truth and foil class presented in the classification experiment for that image. The * indicates the AI classification; thus, first (last) two columns correspond to AI being correct (incorrect).} 
  \label{fig:examples}
\end{figure}

\begin{table*}[t]
\caption{Summary of hypotheses and results. See Section~\ref{sec:statistical_analysis} for details on the hypotheses.}
\label{tab:results}
\vskip 0.15in
\begin{center}
\begin{small}
\begin{tabular}{lllcr}
\toprule
Hypothesis & 
Test & 
P-value \\
\midrule
1) Participants expect the AI to mostly be correct &
Chi-square & 
$< .0001$ \\
2) Participants find explanations to be more helpful when the AI is wrong ($\beta_3 < 0$) &
GLM &
$<. 001$ \\ 
3) Full cognitive model recovers Hypothesis 2 & 
GLM & 
$ .006$ \\
4) Full cognitive model matches participants better than the prior-only model & 
LOO-CV MSE &
$.006$        \\
5) Full cognitive model matches participants better than the L1-norm model &
LOO-CV MSE &
$.02$ \\
6) Full cognitive model matches participants better than the Beta-distribution model &
LOO-CV MSE &
$.003$ \\
Modelled fidelity matches participants' measured fidelity &
Spearman correlation &
$<.0001$ ($\rho=.86$) \\
\bottomrule
\end{tabular}
\end{small}
\end{center}
\vskip -0.1in
\end{table*}

Moreover, model comparisons demonstrate the importance of the psychologically informed components of our cognitive model as predicted in hypotheses 4-6. 
Fourth, the full model predicts aggregate participant responses significantly better than a prior-only model (LOO-CV MSE difference[95\% CI] = 0.014[0.004-0.024], $t$(88) = 2.84, $p$ = .006; see Figure~\ref{fig:results}D), implying that our likelihood function captured explanation-specific belief-updating. Fifth, the full model outperforms the L1-norm model (LOO-CV MSE difference[95\% CI] = 0.005[0.001-0.010], $t$(88) = 2.33, $p$ = .02), implying that the psychological plausibility of the similarity space improves predictive accuracy. Sixth, the full model outperformed the Beta-distribution model (LOO-CV MSE difference[95\% CI] = 0.015[0.005-0.025], $t$(88) = 2.96, $p$ = .003), consistent with the monotonic decay in generalization behavior being present in interpreting XAI explanations. Table~\ref{tab:results} summarizes the hypotheses and statistical results.
 
Comparing Figure~\ref{fig:results}A and B, we note that the cognitive model captured the effect of the explanations over the whole range of explanation quality tested. Firstly, the XAI saliency maps varied in visual quality, with some being more diffuse than others (Figure~\ref{fig:examples}). The variation in quality is also apparent from the wide spectrum of fidelity and in the change in fidelity from the control to the explanation conditions being both positive and negative (mean=8\%, SD=17\%; Figure~\ref{fig:results}A). In particular, some explanations were poor and resulted in a reduction in fidelity (33/89 images). The full model could capture the change in fidelity across all these variations, including explanations of poor quality.

\section{Conclusion}
Systematic, scientific efforts enable the goal of safe, effective, and ethical AI by providing generalizable theories that inform development of effective explanations across domains. 
We have provided a simple, psychologically grounded, quantitative model of human interpretation of explanations for AI systems. 
Our theory proposes that humans reason about AI systems in the same way they reason about human agents: by projecting beliefs.
Human predictions of AI behavior contingent on explanations are based on generalization functions in similarity space, consistent with those identified in cognitive science. 
Our theory is, therefore, general and can be applied and tested across XAI methods and application domains.

\section{Broader impacts}
Most previous attempts at formalizing explainability has focused on general qualities inherent to explanations, but ignored the agents who actively interprets the explanations. For example, XAI methods prefer less complex explanations to not incur a heavy cognitive load, but do not account for the explainee's inferential biases. Here, we have argued for a more specific formalization of explainability that considers and exploits the distinct nature of human belief-updating. To the best of our knowledge, our theory is the first empirically validated, mathematical formulation of how humans infer AI judgements from provided explanations.

The work presented here can be extended in several directions for wider applicability and scalability. First, the task we used to evaluate user understanding focused on predicting AI image classification. This is foundational to --- but a few steps removed from --- real-world applications of XAI such as auditing and debugging \cite{kaur2020interpreting}. Further cognitive and behavioral modelling of inference from explanation in the context of such applications can help bridge this gap. Second, our work focuses on saliency maps, whereas many real-world applications of attribution methods attempt to explain tabular and text data. Our framework can be extended to abstract, non-visual features encoded in tabular and text data by drawing on the large psychological literature on similarity judgements in different domains. Empirical validation of our theory in these contexts requires further research. Third, our framework currently relies on measurements of human generated explanations to predict the interpretability of AI explanations, which limits scalability. If approximations of human generated explanations could be automatically generated, the utility of our theory for solving practical XAI problems would increase.

Lastly, we note three broad implications of a psychological theory of explanation: (1) Such a theory improves the accuracy of explanation by integrating theories from cognitive science to better model human behavior. Having a formal framework that captures how humans interpret explanations allows researchers to develop ML methods that leverage the mathematical template to better reason about and accommodate human partners in human-machine teams. (2) A general theory of human inference from explanations reduces the need for validation experiments by virtue of being reusable across XAI methods. To give an example, our theory can be used to seed experiments validating other saliency methods by indicating explanations that are likely to be successful. (3) A psychological theory of explanation improves understanding of explainability by better integrating the human and machine components of this problem. ML researchers can leverage the constraints in the explanation space highlighted by our theory to design machines that could avoid explanations that are unlikely to be informative to human users.

\section*{Software and Data}


All experiments, mathematical models, analysis code, and hypothesis tests were preregistered (\url{https://osf.io/4n67p}). User study data that support the findings of this study are available upon request.


\section*{Acknowledgements}
We thank Zhaobin Li and Alexander Lowen for prototyping the classification and  drawing experiments, respectively. This material is based on research sponsored by the Air Force Research Laboratory and DARPA under agreement number FA8750-17-2-0146 to P.S. and S.C.-H.Y. The U.S. Government is authorized to reproduce and distribute reprints for Governmental purposes notwithstanding any copyright notation thereon. This work was also supported by NSF MRI 1828528 to P.S.


\bibliography{references}

\begin{thebibliography}{42}
\providecommand{\natexlab}[1]{#1}
\providecommand{\url}[1]{\texttt{#1}}
\expandafter\ifx\csname urlstyle\endcsname\relax
  \providecommand{\doi}[1]{doi: #1}\else
  \providecommand{\doi}{doi: \begingroup \urlstyle{rm}\Url}\fi

\bibitem[Adadi \& Berrada(2018)Adadi and Berrada]{adadi2018peeking}
Adadi, A. and Berrada, M.
\newblock Peeking inside the black-box: a survey on explainable artificial
  intelligence (xai).
\newblock \emph{IEEE access}, 6:\penalty0 52138--52160, 2018.

\bibitem[Adebayo et~al.(2020)Adebayo, Muelly, Liccardi, and
  Kim]{adebayo2020debugging}
Adebayo, J., Muelly, M., Liccardi, I., and Kim, B.
\newblock Debugging tests for model explanations.
\newblock \emph{Advances in Neural Information Processing Systems},
  33:\penalty0 700--712, 2020.

\bibitem[Arrieta et~al.(2020)Arrieta, D{\'\i}az-Rodr{\'\i}guez, Del~Ser,
  Bennetot, Tabik, Barbado, Garc{\'\i}a, Gil-L{\'o}pez, Molina, Benjamins,
  et~al.]{arrieta2020explainable}
Arrieta, A.~B., D{\'\i}az-Rodr{\'\i}guez, N., Del~Ser, J., Bennetot, A., Tabik,
  S., Barbado, A., Garc{\'\i}a, S., Gil-L{\'o}pez, S., Molina, D., Benjamins,
  R., et~al.
\newblock Explainable artificial intelligence ({XAI}): Concepts, taxonomies,
  opportunities and challenges toward responsible ai.
\newblock \emph{Information Fusion}, 58:\penalty0 82--115, 2020.

\bibitem[Blanc et~al.(2021)Blanc, Lange, and Tan]{blanc2021provably}
Blanc, G., Lange, J., and Tan, L.-Y.
\newblock Provably efficient, succinct, and precise explanations.
\newblock \emph{Advances in Neural Information Processing Systems}, 34, 2021.

\bibitem[Buckner \& Carroll(2007)Buckner and Carroll]{buckner2007self}
Buckner, R.~L. and Carroll, D.~C.
\newblock Self-projection and the brain.
\newblock \emph{Trends in cognitive sciences}, 11\penalty0 (2):\penalty0
  49--57, 2007.

\bibitem[Chen et~al.(2021)Chen, Plumb, Topin, and Talwalkar]{chen2021simulated}
Chen, V., Plumb, G., Topin, N., and Talwalkar, A.
\newblock Simulated user studies for explanation evaluation.
\newblock In \emph{{Workshop at NeurIPS2021: eXplainable AI approaches for
  debugging and diagnosis}.}, 2021.

\bibitem[Doshi-Velez \& Kim(2017)Doshi-Velez and Kim]{doshi2017towards}
Doshi-Velez, F. and Kim, B.
\newblock Towards a rigorous science of interpretable machine learning.
\newblock \emph{arXiv preprint arXiv:1702.08608}, 2017.

\bibitem[Gallese \& Goldman(1998)Gallese and Goldman]{gallese1998mirror}
Gallese, V. and Goldman, A.
\newblock Mirror neurons and the simulation theory of mind-reading.
\newblock \emph{Trends in cognitive sciences}, 2\penalty0 (12):\penalty0
  493--501, 1998.

\bibitem[Gogas \& Papadimitriou(2021)Gogas and Papadimitriou]{gogas2021machine}
Gogas, P. and Papadimitriou, T.
\newblock Machine learning in economics and finance.
\newblock \emph{Computational Economics}, 57\penalty0 (1):\penalty0 1--4, 2021.

\bibitem[Goldstone \& Son(2012)Goldstone and Son]{goldstone2012similarity}
Goldstone, R.~L. and Son, J.~Y.
\newblock \emph{Similarity.}
\newblock Oxford University Press, 2012.

\bibitem[Guidotti et~al.(2018)Guidotti, Monreale, Ruggieri, Turini, Giannotti,
  and Pedreschi]{guidotti2018survey}
Guidotti, R., Monreale, A., Ruggieri, S., Turini, F., Giannotti, F., and
  Pedreschi, D.
\newblock A survey of methods for explaining black box models.
\newblock \emph{ACM computing surveys (CSUR)}, 51\penalty0 (5):\penalty0 1--42,
  2018.

\bibitem[Gunning \& Aha(2019)Gunning and Aha]{gunning2019darpa}
Gunning, D. and Aha, D.
\newblock Darpa’s explainable artificial intelligence (xai) program.
\newblock \emph{AI Magazine}, 40\penalty0 (2):\penalty0 44--58, 2019.

\bibitem[Harris et~al.(2018)Harris, Clithero, and
  Hutcherson]{harris2018accounting}
Harris, A., Clithero, J.~A., and Hutcherson, C.~A.
\newblock Accounting for taste: A multi-attribute neurocomputational model
  explains the neural dynamics of choices for self and others.
\newblock \emph{Journal of Neuroscience}, 38\penalty0 (37):\penalty0
  7952--7968, 2018.

\bibitem[Hase \& Bansal(2020)Hase and Bansal]{hase2020evaluating}
Hase, P. and Bansal, M.
\newblock Evaluating explainable ai: Which algorithmic explanations help users
  predict model behavior?
\newblock In \emph{Proceedings of the 58th Annual Meeting of the Association
  for Computational Linguistics}, pp.\  5540--5552, 2020.

\bibitem[He et~al.(2016)He, Zhang, Ren, and Sun]{he2016deep}
He, K., Zhang, X., Ren, S., and Sun, J.
\newblock Deep residual learning for image recognition.
\newblock \emph{Proceedings of the IEEE Conference on Computer Vision and
  Pattern Recognition}, pp.\  770--778, 2016.

\bibitem[Hendrycks et~al.(2021)Hendrycks, Zhao, Basart, Steinhardt, and
  Song]{hendrycks2021natural}
Hendrycks, D., Zhao, K., Basart, S., Steinhardt, J., and Song, D.
\newblock Natural adversarial examples.
\newblock \emph{Proceedings of the IEEE/CVF Conference on Computer Vision and
  Pattern Recognition}, pp.\  15262--15271, 2021.

\bibitem[Hooker et~al.(2019)Hooker, Erhan, Kindermans, and
  Kim]{hooker2019benchmark}
Hooker, S., Erhan, D., Kindermans, P.-J., and Kim, B.
\newblock A benchmark for interpretability methods in deep neural networks.
\newblock \emph{Advances in neural information processing systems}, 32, 2019.

\bibitem[Jesus et~al.(2021)Jesus, Bel{\'e}m, Balayan, Bento, Saleiro, Bizarro,
  and Gama]{jesus2021can}
Jesus, S., Bel{\'e}m, C., Balayan, V., Bento, J., Saleiro, P., Bizarro, P., and
  Gama, J.
\newblock How can i choose an explainer? an application-grounded evaluation of
  post-hoc explanations.
\newblock In \emph{Proceedings of the 2021 ACM Conference on Fairness,
  Accountability, and Transparency}, pp.\  805--815, 2021.

\bibitem[Kaur et~al.(2020)Kaur, Nori, Jenkins, Caruana, Wallach, and
  Wortman~Vaughan]{kaur2020interpreting}
Kaur, H., Nori, H., Jenkins, S., Caruana, R., Wallach, H., and Wortman~Vaughan,
  J.
\newblock Interpreting interpretability: understanding data scientists' use of
  interpretability tools for machine learning.
\newblock In \emph{Proceedings of the 2020 CHI conference on human factors in
  computing systems}, pp.\  1--14, 2020.

\bibitem[Krishna et~al.(2022)Krishna, Han, Gu, Pombra, Jabbari, Wu, and
  Lakkaraju]{krishna2022disagreement}
Krishna, S., Han, T., Gu, A., Pombra, J., Jabbari, S., Wu, S., and Lakkaraju,
  H.
\newblock The disagreement problem in explainable machine learning: A
  practitioner's perspective.
\newblock \emph{arXiv preprint arXiv:2202.01602}, 2022.

\bibitem[Lakkaraju \& Bastani(2020)Lakkaraju and Bastani]{lakkaraju2020fool}
Lakkaraju, H. and Bastani, O.
\newblock " how do i fool you?" manipulating user trust via misleading black
  box explanations.
\newblock In \emph{Proceedings of the AAAI/ACM Conference on AI, Ethics, and
  Society}, pp.\  79--85, 2020.

\bibitem[Li et~al.(2021)Li, Shi, Li, Bai, Cao, and Chen]{li2021experimental}
Li, X.-H., Shi, Y., Li, H., Bai, W., Cao, C.~C., and Chen, L.
\newblock An experimental study of quantitative evaluations on saliency
  methods.
\newblock In \emph{Proceedings of the 27th ACM SIGKDD Conference on Knowledge
  Discovery \& Data Mining}, pp.\  3200--3208, 2021.

\bibitem[Linardatos et~al.(2021)Linardatos, Papastefanopoulos, and
  Kotsiantis]{linardatos2021explainable}
Linardatos, P., Papastefanopoulos, V., and Kotsiantis, S.
\newblock Explainable ai: A review of machine learning interpretability
  methods.
\newblock \emph{Entropy}, 23\penalty0 (1):\penalty0 18, 2021.

\bibitem[Miller(2019)]{miller2019explanation}
Miller, T.
\newblock Explanation in artificial intelligence: Insights from the social
  sciences.
\newblock \emph{Artificial intelligence}, 267:\penalty0 1--38, 2019.

\bibitem[Narayanan et~al.(2018)Narayanan, Chen, He, Kim, Gershman, and
  Doshi-Velez]{narayanan2018humans}
Narayanan, M., Chen, E., He, J., Kim, B., Gershman, S., and Doshi-Velez, F.
\newblock How do humans understand explanations from machine learning systems?
  an evaluation of the human-interpretability of explanation.
\newblock \emph{arXiv preprint arXiv:1802.00682}, 2018.

\bibitem[Nass \& Moon(2000)Nass and Moon]{nass2000machines}
Nass, C. and Moon, Y.
\newblock Machines and mindlessness: Social responses to computers.
\newblock \emph{Journal of social issues}, 56\penalty0 (1):\penalty0 81--103,
  2000.

\bibitem[Nass et~al.(1995)Nass, Moon, Fogg, Reeves, and Dryer]{nass1995can}
Nass, C., Moon, Y., Fogg, B.~J., Reeves, B., and Dryer, D.~C.
\newblock Can computer personalities be human personalities?
\newblock \emph{International Journal of Human-Computer Studies}, 43\penalty0
  (2):\penalty0 223--239, 1995.

\bibitem[Pesapane et~al.(2018)Pesapane, Codari, and
  Sardanelli]{pesapane2018_ai_radiology}
Pesapane, F., Codari, M., and Sardanelli, F.
\newblock Artificial intelligence in medical imaging: threat or opportunity?
  radiologists again at the forefront of innovation in medicine.
\newblock \emph{European radiology experimental}, 2\penalty0 (1):\penalty0
  1--10, 2018.

\bibitem[Petsiuk et~al.(2018)Petsiuk, Das, and Saenko]{petsiuk2018rise}
Petsiuk, V., Das, A., and Saenko, K.
\newblock {RISE: Randomized Input Sampling for Explanation of Black-box
  Models}.
\newblock \emph{29th British Machine Vision Conference}, 2018.

\bibitem[Poursabzi-Sangdeh et~al.(2021)Poursabzi-Sangdeh, Goldstein, Hofman,
  Wortman~Vaughan, and Wallach]{poursabzi2021manipulating}
Poursabzi-Sangdeh, F., Goldstein, D.~G., Hofman, J.~M., Wortman~Vaughan, J.~W.,
  and Wallach, H.
\newblock Manipulating and measuring model interpretability.
\newblock In \emph{Proceedings of the 2021 CHI Conference on Human Factors in
  Computing Systems}, pp.\  1--52, 2021.

\bibitem[Pruthi et~al.(2022)Pruthi, Bansal, Dhingra, Soares, Collins, Lipton,
  Neubig, and Cohen]{pruthi2022evaluating}
Pruthi, D., Bansal, R., Dhingra, B., Soares, L.~B., Collins, M., Lipton, Z.~C.,
  Neubig, G., and Cohen, W.~W.
\newblock Evaluating explanations: How much do explanations from the teacher
  aid students?
\newblock \emph{Transactions of the Association for Computational Linguistics},
  10:\penalty0 359--375, 2022.

\bibitem[Qi et~al.(2020)Qi, Khorram, and Fuxin]{qi2020visualizing}
Qi, Z., Khorram, S., and Fuxin, L.
\newblock Visualizing deep networks by optimizing with integrated gradients.
\newblock In \emph{Proceedings of the AAAI Conference on Artificial
  Intelligence}, volume~34, pp.\  11890--11898, 2020.

\bibitem[Ribeiro et~al.(2016)Ribeiro, Singh, and Guestrin]{ribeiro2016lime}
Ribeiro, M.~T., Singh, S., and Guestrin, C.
\newblock " why should i trust you?" explaining the predictions of any
  classifier.
\newblock In \emph{Proceedings of the 22nd ACM SIGKDD international conference
  on knowledge discovery and data mining}, pp.\  1135--1144, 2016.

\bibitem[Russakovsky et~al.(2015)Russakovsky, Deng, Su, Krause, Satheesh, Ma,
  Huang, Karpathy, Khosla, Bernstein, Berg, and Fei-Fei]{ILSVRC15}
Russakovsky, O., Deng, J., Su, H., Krause, J., Satheesh, S., Ma, S., Huang, Z.,
  Karpathy, A., Khosla, A., Bernstein, M., Berg, A.~C., and Fei-Fei, L.
\newblock {ImageNet Large Scale Visual Recognition Challenge}.
\newblock \emph{International Journal of Computer Vision (IJCV)}, 115\penalty0
  (3):\penalty0 211--252, 2015.
\newblock \doi{10.1007/s11263-015-0816-y}.

\bibitem[Schurz et~al.(2021)Schurz, Radua, Tholen, Maliske, Margulies, Mars,
  Sallet, and Kanske]{schurz2021toward}
Schurz, M., Radua, J., Tholen, M.~G., Maliske, L., Margulies, D.~S., Mars,
  R.~B., Sallet, J., and Kanske, P.
\newblock Toward a hierarchical model of social cognition: A neuroimaging
  meta-analysis and integrative review of empathy and theory of mind.
\newblock \emph{Psychological Bulletin}, 147\penalty0 (3):\penalty0 293, 2021.

\bibitem[Shepard(1987)]{shepard1987toward}
Shepard, R.~N.
\newblock Toward a universal law of generalization for psychological science.
\newblock \emph{Science}, 237\penalty0 (4820):\penalty0 1317--1323, 1987.

\bibitem[Sloman \& Rips(1998)Sloman and Rips]{sloman1998similarity}
Sloman, S.~A. and Rips, L.~J.
\newblock Similarity as an explanatory construct.
\newblock \emph{Cognition}, 65\penalty0 (2-3):\penalty0 87--101, 1998.

\bibitem[Sundararajan et~al.(2017)Sundararajan, Taly, and
  Yan]{sundararajan2017axiomatic}
Sundararajan, M., Taly, A., and Yan, Q.
\newblock Axiomatic attribution for deep networks.
\newblock In \emph{International Conference on Machine Learning}, pp.\
  3319--3328. PMLR, 2017.

\bibitem[Suzuki et~al.(2016)Suzuki, Jensen, Bossaerts, and
  O’Doherty]{suzuki2016behavioral}
Suzuki, S., Jensen, E.~L., Bossaerts, P., and O’Doherty, J.~P.
\newblock Behavioral contagion during learning about another agent’s
  risk-preferences acts on the neural representation of decision-risk.
\newblock \emph{Proceedings of the National Academy of Sciences}, 113\penalty0
  (14):\penalty0 3755--3760, 2016.

\bibitem[Tarantola et~al.(2017)Tarantola, Kumaran, Dayan, and
  De~Martino]{tarantola2017prior}
Tarantola, T., Kumaran, D., Dayan, P., and De~Martino, B.
\newblock Prior preferences beneficially influence social and non-social
  learning.
\newblock \emph{Nature communications}, 8\penalty0 (1):\penalty0 1--14, 2017.

\bibitem[Yang et~al.(2021)Yang, Vong, Sojitra, Folke, and
  Shafto]{yang2021mitigating}
Yang, S. C.-H., Vong, W.~K., Sojitra, R.~B., Folke, T., and Shafto, P.
\newblock Mitigating belief projection in explainable artificial intelligence
  via bayesian teaching.
\newblock \emph{Scientific reports}, 11\penalty0 (1):\penalty0 1--17, 2021.

\bibitem[Yeh et~al.(2019)Yeh, Hsieh, Suggala, Inouye, and
  Ravikumar]{yeh2019fidelity}
Yeh, C.-K., Hsieh, C.-Y., Suggala, A., Inouye, D.~I., and Ravikumar, P.~K.
\newblock On the (in) fidelity and sensitivity of explanations.
\newblock \emph{Advances in Neural Information Processing Systems},
  32:\penalty0 10967--10978, 2019.

\end{thebibliography}
\bibliographystyle{icml2022}

\newpage
\appendix
\onecolumn
\section{Appendix}


\subsection{Saliency map generation through Bayesian Teaching}\label{app:BTmap}

Following previous work \cite{yang2021mitigating}, we generated saliency maps by using Bayesian Teaching to select pixels of an image that help a learner model to arrive at the targeted prediction. Let $q_{teacher}(\mask  \mid  c, \image)$ be the probability that a mask $\mask$ will lead the learner model to predict the image $\image$ to be in class $c$ when the mask is applied to the image. This is expressed by Bayes' rule as
 \begin{align*}
   q_{teacher}(\mask \mid c, \image) = \frac{Q_{learner}(c \mid \image, \mask ) p(\mask )}
   {\int_{\Omega_\mathbf{M}} Q_{learner}(c \mid \image, \mask') p(\mask')}.
 \end{align*}
Here, $Q_{learner}(c  \mid  \image, \mask)$ is the probability that the ResNet-50 model with pre-trained ImageNet weights will predict the $\image$ masked by $\mask$ to be $c$; $p(\mask)$ is the prior probability of $\mask$; and $\Omega_\mathbf{M} = [0, 1]^{W \times H}$ is the space of all possible masks on an image with $W\times H$ pixels. We used a sigmoid-function squashed Gaussian process prior for $p(\mask)$.

Instead of sampling the saliency maps directly from the above equation, we find the expected saliency map for each image by Monte Carlo integration:
 \begin{align}
 \text{E}[\mathbf{M}  \mid  \image,c]
 &=\int_{\Omega_\mathbf{M}} \mask\ q_{teacher}(\mask \mid c, \image) \nonumber\\
 &\approx \frac{\sum_{i=1}^N \mask_i\ Q_{learner}(c \mid \image, \mask_i)}{\sum_{i=1}^N Q_{learner}(c \mid \image, \mask_i)},
 \label{eq:map}
 \end{align}
where $\mask_i$ are samples from the prior distribution $p(\mask)$, and $N=1000$ is the number of Monte Carlo samples used. The expected mask is used as the saliency map explanation $\ex$.

To generate the saliency map $\ex$ for an image $\image$, we first resized $\image$ to be 224-by-224 pixels. A set of 1000 2D functions were sampled from a 2D Gaussian process (GP) with an overall variance of $100$, a constant mean of $-100$, and a radial-basis-function kernel with length scale 22.4 pixels in both dimensions. The sampled functions were evaluated on a 224-by-224 grid, and the function values were mostly in the range of $[-500,300]$. A sigmoid function, $1 / (1 + \exp(-a))$, was applied to the sampled functions to transform each of the function values $a$ to be within the range $[0,1]$. This resulted in 1000 masks. The mean of the GP controlled how many effective zeros there were in the mask, and the variance of the GP determined how fast neighboring pixel values in the mask changed from zero to one. The 1000 masks were the $\mask_i$'s in Equation~\ref{eq:map}. We produced 1000 masked images by element-wise multiplying the image $\image$ with each of the masks. The term $Q_{learner}(c  \mid  \image, \mask_i)$ was the ResNet-50's predictive probability that the $i^\textrm{th}$ masked image was in class $c$. Having obtained these predictive probabilities, we averaged the 1000 masks according to Equation~\ref{eq:map} to produce the saliency map $\ex$ of image $\image$.

\subsection{Similarity calculations}\label{app:sim}
The likelihood function of our cognitive model takes as input the similarity between the observed AI saliency map and the image-level aggregates of the participant-generated maps for the ground truth and foil classes. Therefore, before we explain the similarity calculations, we discuss the mask aggregation and scaling. For the aggregation, we compute the pixel-level mean across all participants who passed the exclusion thresholds outlined above. For the Sloman similarity \cite{sloman1998similarity}, we then min-max scale both the observed AI saliency maps and the aggregated participant-generated masks, so that the highest pixel value in each map/mask is 1, and the lowest pixel value is 0. For the L1 norm, we scale the observed maps and participant-generated masks so that all pixel-values for each map/mask sums to one. We compute separate similarity measures by comparing each of the observed AI saliency maps with the participant-generated masks for the ground-truth label and the foil label.

The Sloman similarity captures the overlap between features (regions important for the classification) in the observed AI saliency map and those in the participant-generated masks \cite{sloman1998similarity}. Specifically, the Sloman similarity takes this form:
\begin{align*}
sim\!\left[\ex(c,\image) \,,\, \ex^{P}\!(c,\image) \right] = \frac{\sum_i \ex_i(c,\image) \ex^{P}_i\!(c,\image)}
{ \sqrt{ \sum_i {\ex_i(c,\image)}^2} \sqrt{ \sum_i {\ex^{P}_i\!(c,\image)}^2} },
\end{align*}
where $\ex$ is the observed AI saliency map, and $\ex^{P}$ is the aggregated human-generated mask, which is referred to as the projected map in the main text. The $i$ indexes the pixels in the map/mask. The numerator captures feature overlap as the intersection between highly salient regions. The denominator normalizes this area of intersection to be between [0,1].

We compute the L1 norm dissimilarity by summing the absolute pixel-wise difference between the observed AI saliency map and the participant-generated mask:
\begin{align*}
\sum_i \lvert \ex_i(c,\image) - \ex^{P}_i\!(c,\image) \rvert,
\end{align*}
where $c$ indexes the ground-truth or foil class.

\subsection{Detailed statistical results}\label{app:results}
\subsubsection{Hypothesis 1}
In line with our preregistered hypothesis, participants believed that the AI would correctly classify the image in 73\% of the trials, which is significantly different from chance according to a chi-squared test ($\chi^{2}$ = 802.28, $p$ $<$ .0001).

\subsubsection{Hypothesis 2}
In our preregistration we hypothesized that $\beta_{2}$ would be significantly positive, and that $\beta_{3}$ would be significantly negative. Both of these hypotheses were supported, meaning that we found that explanations improved the fidelity between participant responses and AI classifications when the AI makes a mistake ($\beta_{2}$ = 0.14, SE = 0.03, $t$ = 4.69, $p$ $<$ .0001), and that the impact of explanations on fidelity was reduced when the AI is correct ($\beta_{3}$ = -0.17, SE = 0.05, $t$ = -3.40, $p$ $<$ .001). See Table~\ref{table:empirical_regression}.

\begin{table}[h!]
\begin{center}
\begin{tabular}{l l l}
\hline
Label & Parameter & Coefficient (SE) \\
\hline
Intercept & $\beta_{0}$ & $0.33^{***}(0.02)$  \\
AICorrect & $\beta_{1}$ & $0.53^{***}(0.04)$  \\
ExplanationCondition & $\beta_{2}$ & $0.14^{***}(0.03)$  \\
AICorrect $\times$ ExplanationCondition & $\beta_{3}$ & $-0.17^{***}(0.05)$ \\
\hline
R$^2$      & & $0.65$ \\
Adj. R$^2$ & & $0.64$ \\
$I$   & & $178$ \\
\hline
\multicolumn{2}{l}{\scriptsize{$^{***}p<0.001$; $^{**}p<0.01$; $^{*}p<0.05$}}
\end{tabular}
\caption{Image-level empirical explanation results. AICorrect and ExplanationCondition correspond to $A_i$ and $E_i$ in Equation~\ref{eq:lm}, respectively.}
\label{table:empirical_regression}
\end{center}
\end{table}

\subsubsection{Hypothesis 3}
Our preregistered hypothesis was that these results should match the empirical results in the explanation condition. Specifically, $\beta_{2}$ should be significantly positive, and $\beta_{3}$ should be significantly negative. As hypothesized, the model posteriors recover the effect of explanations on fidelity during mistake trials ($\beta_{2}$ = 0.12, SE = 0.03, $t$ = 3.90, $p$ $<$ .001), as well as the moderation effect between explanations and AI correctness ($\beta_{3}$ = -0.14, SE = 0.05, $t$ = -2.77, $p$ = .006). See Table~\ref{table:model_regression}.

\begin{table}[h!]
\begin{center}
\begin{tabular}{l l l}
\hline
Label & Parameter & Coefficient (SE) \\
\hline
Intercept & $\beta_{0}$ & $0.33^{***}(0.02)$  \\
AICorrect & $\beta_{1}$ & $0.53^{***}(0.04)$  \\
ExplanationCondition & $\beta_{2}$ & $0.12^{***}(0.03)$  \\
AICorrect $\times$ ExplanationCondition & $\beta_{3}$ & $-0.14^{**}(0.05)$ \\
\hline
R$^2$      & & $0.65$ \\
Adj. R$^2$ & & $0.65$ \\
$I$   & & $178$ \\
\hline
\multicolumn{2}{l}{\scriptsize{$^{***}p<0.001$; $^{**}p<0.01$; $^{*}p<0.05$}}
\end{tabular}
\caption{Image-level modelled explanation results. AICorrect and ExplanationCondition correspond to $A_i$ and $E_i$ in Equation~\ref{eq:lm}, respectively.}
\label{table:model_regression}
\end{center}
\end{table}

\subsubsection{Hypothesis 4}
In line with our hypothesis, the full model significantly outperforms a prior-only model (LOO-CV MSE difference [95\% CI] = 0.014[0.004-0.024], $t$(88) = 2.84, $p$ = .006). 

\subsubsection{Hypothesis 5}
In line with our hypothesis, our full model outperforms the L1-norm model (LOO-CV MSE difference [95\% CI] = 0.005[0.001-0.010], $t$(88) = 2.33, $p$ = .02).

\subsubsection{Hypothesis 6}
In line with our hypothesis, our full model outperformed the Beta-distribution model (LOO-CV MSE difference [95\% CI] = 0.015[0.005-0.025], $t$(88) = 2.96, $p$ = .003), suggesting that the monotonic decay in generalization behavior is also observed in interpreting XAI explanations.

\subsubsection{Not preregistered analysis}
The Spearman correlation between fidelity based on the empirical data and fidelity based on the model posteriors was statistically significant (Spearman's $\rho$ = .86, $p$ $<$ .0001).

\subsection{Statistical analysis: Pre-registered exclusions}\label{app:exclusions}

Upon completing data collection we learned that our exclusion criterion ($Z > 1.2$) for the drawing experiment data had been too strict, leading us to exclude 82\% of participants for the ground-truth condition and 65\% of participants for the foil condition. We adapted a more generous threshold ($Z > 1.5$) that allowed us to meet our sample size targets and report the results from these larger samples in the main text.  

Below are the results of the analyses on only the data that met the preregistered inclusion threshold. Because the inclusion criteria only differed for the drawing experiment, the results for Hypothesis 1 and 2 are exactly the same as the previous section.

\subsubsection{Hypothesis 3}

See Table~\ref{table:model_regression_preregistered}.
\begin{table}[h!]
\begin{center}
\begin{tabular}{l l l}
\hline
Label & Parameter & Coefficient (SE) \\
\hline
Intercept & $\beta_{0}$ & $0.33^{***}(0.02)$  \\
AI Correct & $\beta_{1}$ & $0.53^{***}(0.04)$  \\
Explanation Condition & $\beta_{2}$ & $0.11^{***}(0.03)$  \\
AI Correct $\times$ Explanation Condition & $\beta_{3}$ & $-0.13^{*}(0.05)$ \\
\hline
R$^2$      & & $0.65$ \\
Adj. R$^2$ & & $0.65$ \\
$I$   & & $178$ \\
\hline
\multicolumn{2}{l}{\scriptsize{$^{***}p<0.001$; $^{**}p<0.01$; $^{*}p<0.05$}}
\end{tabular}
\caption{Image-level modelled explanation results based on preregistered exclusion thresholds. AICorrect and ExplanationCondition correspond to $A_i$ and $E_i$ in Equation~\ref{eq:lm}, respectively.}
\label{table:model_regression_preregistered}
\end{center}
\end{table}

\subsubsection{Hypothesis 4}
MSE difference [95\% CI] = 0.014[0.005-0.024], $t$(88) = 2.98, $p$ = .003.

\subsubsection{Hypothesis 5}
MSE difference [95\% CI] = 0.007[0.001-0.014], $t$(88) = 2.38, $p$ = .02.

\subsubsection{Hypothesis 6}
MSE difference [95\% CI] = 0.014[0.004-0.024], $t$(88) = 2.85, $p$ = .005.

\subsubsection{Not preregistered analysis}
Spearman correlation at the trial level between empirical fidelity for the explanation condition and posterior fidelity of the cognitive model (Spearman's $\rho$ = .86, $p$ $<$ .0001).


\end{document}